\documentclass[11pt,a4paper]{article}
\usepackage[hyperref]{acl2020}
\usepackage{times}
\usepackage{latexsym}

\usepackage{color}

\usepackage{microtype}

\aclfinalcopy % Uncomment this line for the final submission
\usepackage{multirow}
\usepackage{dblfloatfix}
\usepackage[para,online,flushleft]{threeparttable}
\usepackage{booktabs}
\usepackage{siunitx}
\usepackage{hyperref}
\usepackage{float}
\usepackage{todonotes}
\title{Better Distractions: Transformer-based Distractor Generation\\ and Multiple Choice Question Filtering}

\author{Jeroen Offerijns, Suzan Verberne, Tessa Verhoef \\
  Leiden Institute of Advanced Computer Science, Leiden University}
\date{}

\begin{document}
\maketitle
\begin{abstract}
  For the field of education, being able to generate semantically correct and  educationally  relevant multiple choice questions (MCQs) could have a large impact. While question generation itself is an active research topic, generating distractors (the incorrect multiple choice options) receives much less attention. A missed opportunity, since there is still a lot of room for improvement in this area. In this work, we train a GPT-2 language model to generate three distractors for a given question and text context, using the RACE dataset. Next, we train a BERT language model to answer MCQs, and use this model as a filter, to select only questions that can be answered and therefore presumably make sense. To evaluate our work, we start by using text generation metrics, which show that our model outperforms earlier work on distractor generation (DG) and achieves state-of-the-art performance. Also, by calculating the question answering ability, we show that larger base models lead to better performance. Moreover, we conducted a human evaluation study, which confirmed the quality of the generated questions, but showed no statistically significant effect of the QA filter.
\end{abstract}

\section{Introduction}
\label{introduction}
Over the last two years, Transformer-based language models have progressed from initial development to being adopted in all parts of natural language processing (NLP). This started with ULMFiT \citep{umlfit} and BERT \citep{bert}, which showed the potential of pre-training a large neural network using unsupervised learning. After pre-training, these neural networks can be fine-tuned on specific tasks. During fine-tuning, the weights of the model are tweaked to perform well on a specific task, building upon the knowledge learned during pre-training. This has led to substantial improvements in the state of the art for tasks such as sentiment classification, question answering, and many others. When GPT-2 \citep{gpt2} was released, a huge improvement in text generation ability was obtained. The performance has even been shown to continue to improve with an increase in the size of the language models, ranging from 117M parameters for the smallest GPT-2 model, to 175B parameters for the largest of the GPT-3 \citep{gpt3} models.

\begin{figure*}
\centering
\includegraphics[width=1.0\linewidth]{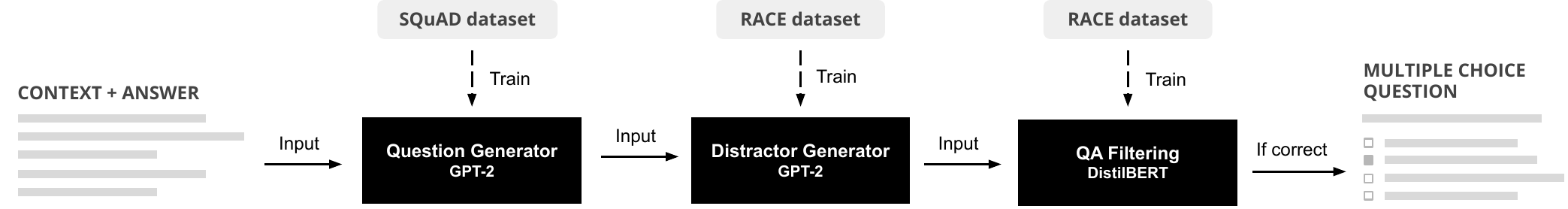}
\caption{Overview of the model architecture: we take a context and answer as input, generate a question, generate three distractors, and use the QA model to filter only correctly answered questions. The distractor and question generator models are based on GPT-2, which is ideal for text generation, while the QA filtering model is based on BERT, which is better for classification problems.}
\label{fig:overview}
\end{figure*}

Within natural language processing, question answering (QA) is a heavily researched field, while the inverse task receives much less attention: question generation (QG) \cite{pan2019recent}. For education, being able to generate semantically correct and educationally relevant questions is a challenging task with clear applications. Yet most of the work in this field focuses on QG for generating synthetic datasets for question answering, rather than seeing it as a goal on its own. For this reason, these papers tend to concentrate only on the task of generating a question from a given context and answer, while the other elements required for multiple choice questions (MCQs) receive much less attention. These elements include selecting the answer and generating the incorrect answers. It is this last part that we focus on: generating incorrect answers, also known as distractors.

For the distractor generation task, we use the RACE dataset \cite{race_dataset}, which contains almost $100,000$ questions. Each of these questions is paired with a context of a single paragraph, the correct answer, and three distractors. We use this to create a distractor generation model, which gives us the ability to generate complete multiple choice questions. Previous work on distractor generation with the RACE dataset \citet{gao2019distractor,zhou2019co} use sequence-to-sequence models to generate the distractors, which leads to low-quality text. The generation of complete multiple choice questions opens up other possibilities, including the ability to create a QA model which chooses the correct answer from four options. We will investigate whether such a multiple choice QA model can be used to filter only correctly answered questions in order to improve the overall quality of question generation models.

The key contributions of this work are:
\begin{itemize}
\item{We fine-tuned a GPT-2 language model for distractor generation on the RACE dataset.}
\item{We fine-tuned a BERT language model for multiple choice question answering on the RACE dataset.}
\item{We proposed a new QA filtering method for improving QG results, by filtering using a multiple choice QA model.}
\end{itemize}

\section{Related work}
\label{relatedwork}

\paragraph{Question generation}
Early question generation models were mainly rule-based: defining patterns of word types and using these to extract phrases from the text, which would be transformed into questions \citep{mitkov_ha_2003_computer, chen_etal_2006_fast,heilman2011automatic}. In the last decade, these rule-based models were mostly replaced by neural networks, primarily sequence-to-sequence architectures \citep{du2017learning,Kim_2019}. However, in the last year, these again are being replaced, now with Transformer-based language models.

The first of such works used BERT to generate questions \citep{Alberti_2019}.
By now, GPT-2 \citep{gpt2} has mostly replaced BERT for QG tasks \citep{klein2019learning,Liu_2020,cho2019contrastive,lopez2020transformerbased}. GPT-2 is a better text generator overall \cite{Wang_2019} due to it being trained solely in a left-to-right fashion, predicting the next word in a sequence of words. This is in contrast with bidirectional models such as BERT, which are trained primarily by predicting masked words. Such masked language modeling training leads to better performance on many NLP classification tasks, due to the bidirectional nature, but is worse at the specific task of text generation.

\paragraph{Distractor generation}
Several previous solutions for distractor generation (DG) are actually ranking models. These include the work by  \citet{liang_etal_2018_distractor}, which ranks distractors from a given candidate set using both feature-based and neural network-based ranking models, and \citet{ren2020knowledgedriven}, who use a knowledge base to generate a distractor candidate set and a learning-to-rank model for selecting distractors.

In 2017, the English language RACE dataset \citep{race_dataset} was published. This was the first large dataset to include distractors along with the questions. Several papers since then have used this to create distractor generation models, including \citet{gao2019distractor}, which used a hierarchical encoder-decoder model with attention to generate distractors. \citet{zhou2019co} improved upon this model by adding co-attention layers and using more tricks to gain better performance. Our works uses Transformer-based language models instead, leading to higher quality outputs.

\paragraph{Multiple choice QA}
The original RACE paper used several models to establish baselines on the multiple choice QA task. Their Gated AR model achieved an accuracy of 44.1\%, which showed the limitations of the models available at that time of publication (2017) for such a complex dataset. Recently, language models have been able to greatly surpass this accuracy, with BERT achieving an accuracy of 73.9\% \citep{lan2019albert}, and the largest variant of ALBERT \citep{lan2019albert} even achieving an accuracy of 82.3\%. We use these advances in question answering models to create a multiple choice QA model and employ this in the context of distractor generation.

\paragraph{QA filtering}
\citet{Alberti_2019} introduced the concept of QA filtering to the domain of question generation. They generate a question, then answer that question using an extractive text QA model. Only when the QA model generates the correct answer, they keep it. This is to ensure roundtrip consistency. \citet{Liu_2020} also used a similar filtering method, but with the explicit goal of generating human-like questions. These approaches differ from our method since we do not generate a textual answer, but we check whether a multiple choice QA model can choose the correct option out of four answers.

\section{Method}
\label{method}
Our system consists of three separate models: a question generator, a distractor generator, and a QA filter. We will outline how we created and trained these models separately, and then we will explain how we used these jointly to improve the overall results. Figure \ref{fig:overview} provides a high-level overview of our complete architecture.

\subsection{Question generation}
While question generation is not the goal of our research, we do use it as input for the other two models. It is used to evaluate the ability of the QA model to filter generated question—answer—distractor tuples. Similar to many recent works \citep{klein2019learning,Liu_2020,lopez2020transformerbased}, we decided to fine-tune a GPT-2 model, in particular the ``small" variant with 117 million parameters. For this task, we used the English SQuAD dataset \citep{squad}, specifically  the training dataset of SQuAD v2. We used SQuAD rather than the RACE dataset for this task, in order to create a model which is similar to most recent works in question generation, which almost exclusively use the SQuAD dataset. We remove questions which are highlighted as being impossible to answer (as specified by humans when the dataset was created), because we want our model to generate answerable questions. After removing these, $86,821$ questions remained.

\begin{figure}[H]
\centering
\includegraphics[width=1.0\linewidth]{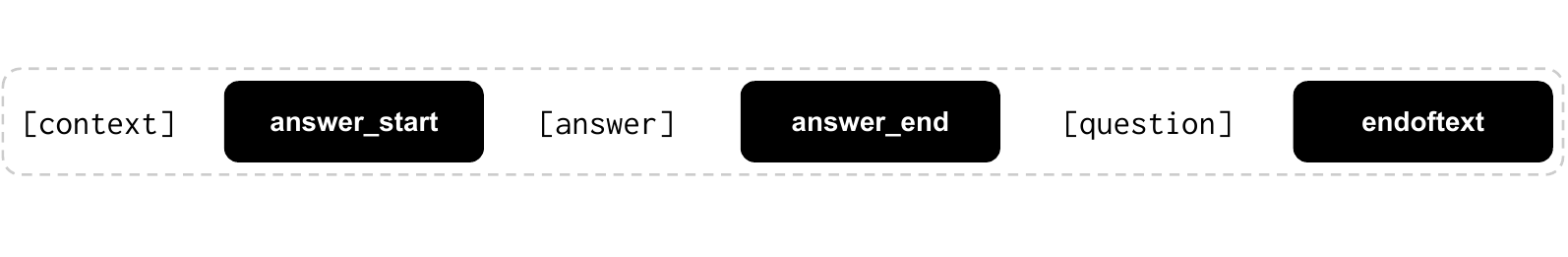}
\caption{Format of the input to the QG model. The black boxes denote special tokens supplied to the tokenizer.}
\label{fig:input_qg}
\end{figure}

We extract context—answer—question tuples from the SQuAD dataset, and tokenize these using the Byte-Pair-Encoding (BPE) tokenizer \citep{sennrich_etal_2016_neural} that GPT-2 uses. Since GPT-2 is a model that learns to generate the next word after a sequence of words, we use special tokens to identify the segments of the inputs. This forces the model to learn to generate the correct elements. The input format is shown in Figure \ref{fig:input_qg}.

This model was implemented in PyTorch \citep{pytorch} using the Transformers library \citep{wolf2019huggingfaces}. The model was already pre-trained by OpenAI on a large text corpus, and we fine-tuned it on our dataset. It was fine-tuned for 3 epochs on the full dataset, using a batch size of 4. The Adam optimizer \citep{adam} was used with a learning rate of \num{5e-5} and an epsilon value of \num{1e-8}. This optimizer improves upon classical stochastic gradient descent by using first and second moments of the gradients to speed up convergence. Using the Adam optimizer is standard practice for Transformer-based models. The learning rate and epsilon values are based on recommendations from \citet{wolf2019huggingfaces}.

% \todo[inline]{Top k filtering, beam search, etc.} %

\subsection{Distractor generation}
\begin{figure}[H]
\centering
\includegraphics[width=1.0\linewidth]{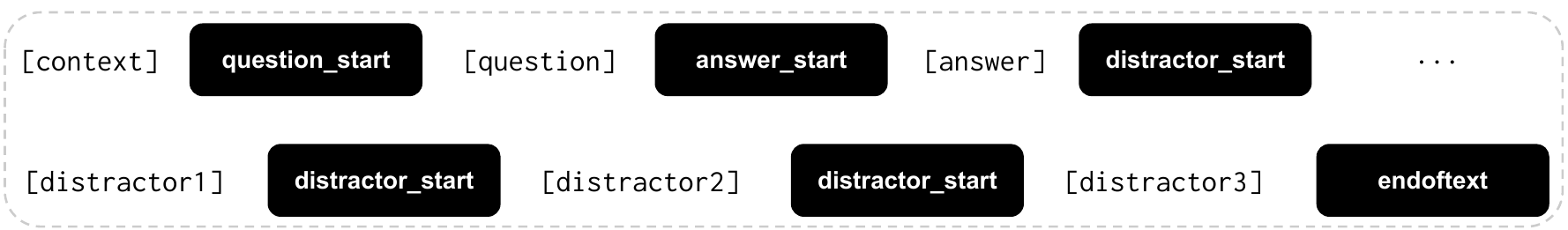}
\caption{Format of the input to the DG model.}
\label{fig:input_dg}
\end{figure}

Similar to the question generation model, we again fine-tune GPT-2, but this time to generate distractors. Since the SQuAD dataset does not contain distractors, we used the RACE dataset \citep{race_dataset} for this model. We do not do any filtering, so we use the full training dataset of $87,866$ questions. We provide the context, question, and answer as input. The context is where the model can draw stylistic influence and semantic information from. The question is what the distractors should be written in relation to. And finally, the answer should be used to make sure that the distractors are different from the answer. The input format is shown in Figure \ref{fig:input_dg}.

This is again tokenized using the BPE tokenizer, and we train the model with the same settings. However, besides training the small GPT-2 model, we also train another model based on the medium GPT-2 variant, with 355 million parameters. We keep the settings the same, except for the batch size which we reduce to 1, since we are limited by the memory usage. %Ideally, we would have also trained the large or extra large GPT-2 variant, but this was not possible%\footnote{Although it is technically possible with the use of gradient checkpointing, we did not get this to work.} 
%on the RTX 2080 TI GPUs that we used.

During generation, we also apply a repetition penalty, as proposed by the authors of the CTRL language model \citep{keskar2019ctrl}. This penalizes the model for generating similar texts, which helps to generate syntactically dissimilar distractors. Moreover, we noticed that the model could sometimes generate less than three distractors, generate non-unique distractors, or generate empty strings as distractors. To alleviate this, we decided to filter non-unique and empty distractors, and to repeat the generation step until three unique and non-empty distractors were found.

\subsection{QA filtering}
\begin{figure}[H]
\centering
\includegraphics[width=1.0\linewidth]{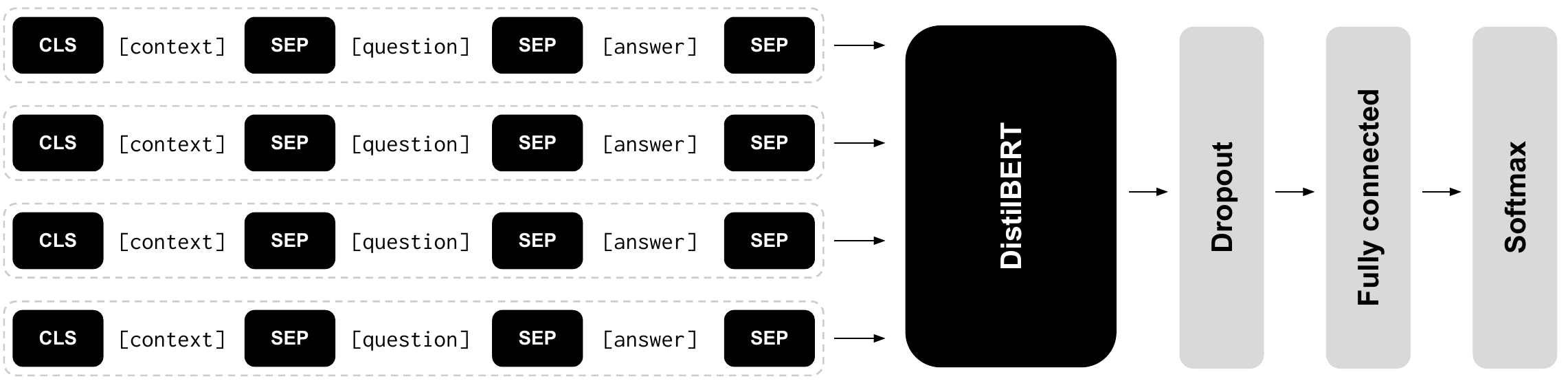}
\caption{Overview of the input and architecture of the QA filtering model. We feed each distractor separately into the DistilBERT model, then use the four outputs to determine the answer.}
\label{fig:qaf_model}
\end{figure}

In order to be able to filter multiple choice questions, we need to have a model which can answer them. To create this, we decided to fine-tune the DistilBERT model \citep{sanh2019distilbert}, with 66 million parameters. This is a distilled version of BERT, retaining 97\% of the performance of the small BERT model, with 40\% less parameters. Most QA research focuses on extractive QA: models where the output is a string, which is extracted from the source document. In our case, we want a QA model which chooses one of the multiple choice options as the correct answer. To accomplish this, we feed context—question—answer tuples into BERT. We then combine the four outputs and feed it through a dropout layer \citep{dropout} for regularization, a fully connected layer for classification, and finally a softmax layer in order to model it as a multi-class classification problem. The input format and the model architecture is shown in Figure \ref{fig:qaf_model}.

This model was trained for 3 epochs, with a fully connected layer dimension of 768, a dropout ratio of 10\%, a batch size of 2, and 8 gradient accumulation steps per batch.\footnote{This simulates a larger batch size, which is required for good performance with a QA model on the RACE dataset \citep{liu2019roberta}.} Again, the Adam optimizer was used,  with a learning rate of \num{3e-5} and an epsilon value of \num{1e-8}.

Once we have the multiple choice QA model, we can use it to filter question—answer—distractor tuples. The intuition behind this QA filter is that when a multiple choice QA model is given perfect information, it should almost always be able to answer a generated question correctly. If not, there could be two type of errors:  either (A) the QA model does not have the capability to answer it,  or (B) the question or distractors are somehow incorrect (i.e. this is a bad question). As for the type A errors, this should be unlikely because the model receives the exact context which is needed to answer the question. Imagine if you had a test and the students would be provided the paragraph which contained the answer for the question right next to every question: students would surely receive high grades. Moreover, QA models have already surpassed human performance on the SQuAD dataset \cite{zhang2020retrospective} and are nearing human performance on the RACE dataset \cite{lan2019albert}, further decreasing the chance of type A errors. Type B errors are exactly what the QA model aims to filter. Therefore, whether the QA model can answer the question should be a good filter for high-quality questions.

\section{Results}
\begin{table*}[t]
\centering
\begin{tabular}{llllll}
\toprule
 & BLEU-1         & BLEU-2         & BLEU-3        & BLEU-4        & ROUGE-L\\
\midrule
\textbf{Dataset questions}\\
\textsc{Seq2seq} \citep{gao2019distractor}           & 25.25          & 11.99          & 6.54          & 3.92          & 13.34          \\
\textsc{HSA} \citep{gao2019distractor}  & 26.93          & 13.57          & 8.00          & 5.21          & 14.45          \\
\textsc{CHN} \citep{zhou2019co} & 27.53          & 13.80          & 8.46          & 5.80          & \textbf{15.11}          \\
\textsc{GPT-2 small} & 60.12 & \textbf{26.56} & \textbf{13.64} & \textbf{9.17} & 12.36 \\
\textsc{GPT-2 medium} & \textbf{60.85} & 26.52 & 13.20 & 8.70 & 12.01                \\
\textsc{GPT-2 medium} (after QA filtering)      & 60.21 & 26.38 & 13.29 & 8.84 & 12.00          \\
\midrule
\textbf{Generated questions}\\
\textsc{GPT-2 small} & 57.08 & 24.14 & 11.73 & 7.59 & 10.40                \\
\textsc{GPT-2 medium} & 57.66 & 24.00 & 11.29 & 7.14 & 9.79                \\
\textsc{GPT-2 medium} (after QA filtering)      & 56.60 & 23.50 & 11.05 & 7.03 & 9.69          \\
\bottomrule
\end{tabular}
\caption{Text generation quality of the distractor generation model. The DG scores are calculated separately for each distractor, and then averaged over all three distractors.}
\label{table:genquality}
\end{table*}

To evaluate our work, we used three different approaches: evaluating the text generation quality using standardized metrics, evaluating the ability for the QA model to answer the generated questions, and using a human evaluation to complement these two automatic metrics with a human perspective.

\subsection{Quantitative evaluation}
We compare our models against three baselines: the basic sequence-to-sequence distractor generator model from \citet{gao2019distractor}, the improved hierarchical encoder-decoder model with static attention (\textsc{HSA}) from \citet{gao2019distractor}, and the hierarchical model enhanced with co-attention (\textsc{CHN}) from \citet{zhou2019co}. 

\subsubsection{Text generation quality}
As a high-level overview, we use several metrics to calculate the quality of the generated distractors. Specifically, we use the BLEU metric, which uses modified\footnote{BLEU's modified version of precision accounts for overgeneration of words by clipping based on the maximum reference word count.} precision of n-grams to determine the correspondence to human-written text; and we use the ROUGE-L metric, which looks at the longest common subsequence and is a measure of recall. The results of this evaluation can be found in Table \ref{table:genquality}. By default, we use questions from the dataset as input to the distractor generator. As a comparison, we also show the case where we are generating the questions as well, to show what the impact is on the results of the distractor generator. This should show lower text generation quality scores, since the question generator will at times generate low quality questions, which would make it harder for the distractor generator to generate high quality distractors.

The distractors in the RACE dataset are on average 5.7 words long, with a standard deviation of 3.3.  This means that for evaluating distractors, the BLEU-1 and BLEU-2 scores are more relevant than BLEU-3 and BLEU-4, since 3-grams and 4-grams occur much less.

Looking at the quantitative results in Table \ref{table:genquality}, the BLEU scores are substantially higher than those reported in previous work. This is in line with what other studies have shown with the use of Transformer-based language models for text generation: these are much better at generating coherent text than previous sequence-to-sequence model based approaches were. However, interestingly, the ROUGE-L score is actually slightly lower than the ROUGE-L scores of prior work. While the BLEU score is a measure of precision, ROUGE-L is a measure of recall. ROUGE measures how many words in the human references appear in the generated distractors.
% A potential explanation for this difference in scores is that the GPT-2 based models are more creative and therefore diverge further from the word distribution of the references than previous models do. This can lead to relatively lower recall.

When looking at the differences between our own models, these seem to be relatively minor. The larger \textsc{GPT-2 medium} model, which has twice the number of parameters as the \textsc{GPT-2 small} model, only gains less than a percentage point (when looking at BLEU-1). This minor change is likely due to the dataset size: the small model is already able to model the distribution well and can already learn to generate distractors like the outputs from the dataset. Furthermore, it appears that only rating distractors after the QA filtering step does not lead to better results. Lastly, the scores for when we generate questions are on average several percentage points lower than when we use questions from the dataset. This makes sense: the question generator will occasionally generate incoherent questions, which will complicate the work of the distractor generator, and lead to outputs which differ more from the reference dataset. But it is worth mentioning that the scores for generated questions are not dramatically lower, which means that the solution for distractor generation proposed here seems to generalise well to the harder task of end-to-end multiple choice question and distractor generation.

\subsubsection{Question answering ability}
\begin{table*}[t]%% {table*}[t] %%
\centering
\begin{tabular}{llllllll}
\toprule
& Dataset questions & Generated questions        \\
\midrule
\textsc{GPT-2 small}          & 51.15\% & 54.29\%         \\
\textsc{GPT-2 medium}  & \textbf{53.36\%} & \textbf{55.90\%}   \\
\bottomrule
\end{tabular}
\caption{Accuracy of the QA model for generated distractors by both DG models.}
\label{table:qa_ability}
\end{table*}

As a second quantitative evaluation, we decided to measure the number of questions answered correctly by the QA model, when the distractors are generated by our model. The better the distractors, the higher this percentage should be, as good distractors should be clearly incorrect answers to the QA model, given the fact that the model has full access to the context. However, as previously noted, the error rate of the QA model is a summation of two errors: errors due to bad distractors or questions, as well as errors made by the QA model itself due to other reasons. Therefore, the accuracy on its own is not meaningful to evaluate the distractors, but it is meaningful as a relative number to compare models.

For the results, see Table \ref{table:qa_ability}. We compare the \textsc{GPT-2 small} and \textsc{medium} models. Again, we also compare the case for which we generate the questions with our question generator, with the case where we use the questions provided by the dataset and only generate the distractors. We can clearly see that using \textsc{GPT-2 medium} for distractor generation, which has twice the number of parameters as \textsc{GPT-2 small}, results in more accurate question answering than the smaller model. Interestingly, the scores when the the questions are also generated, are better than when the questions are taken from the dataset. An explanation for this could be that the question generation model generated questions which are simpler for the QA model to answer, thus leading to higher QA model accuracy.

\subsection{Human evaluation}
Metrics such as BLEU and ROUGE are based merely on comparing text similarity to reference sentences and are therefore limited in their ability to measure the quality of generated text as a human would \citep{bleu_limitations}. Good distractors could definitely be different from the references, which is not accounted for in the text generation quality metrics. Moreover, the text generation quality is calculated on a per-distractor basis, i.e. the first generated distractor for a question is compared with the first reference distractor, and so on. We would argue it makes more sense to compare a generated distractor with all three reference distractors, but we chose this type of comparison to follow the same evaluation methodology as previous works \citep{gao2019distractor,zhou2019co}. Simply reordering distractors would lead to lower text generation quality scores. To account for these limitations in the quantitative evaluation methods, we decided to run a human evaluation. Specifically, we wanted to test the ability of the QA filtering model to filter high quality questions which are answerable by a human. We set up a human evaluation with 4 assessors, each rating 100 generated questions (leading to a total of 310 assessed questions) with the following questions:

\begin{enumerate}
    \item \textbf{Is the question well-formed and can you understand the meaning?} Possible answers include ``Both understandable and well-formed", ``Understandable, but not well-formed.", and ``Neither".
    \item \textbf{If the question is at least understandable, does the answer make sense in relation to the question?} This is a yes, no, or I don't know question.
\end{enumerate}

\begin{figure*}[t]
\centering
\includegraphics[width=0.8\linewidth]{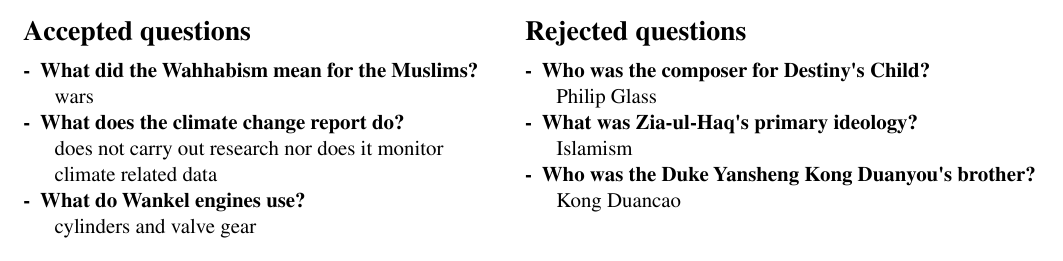}
\caption{Randomly chosen examples of generated questions used for the human evaluation. The difference in quality appears small, which aligns with the results of the human evaluation.}
\label{fig:samples}
\end{figure*}

These questions are based on work done by \citet{Liu_2020}, but we removed the relevancy question since it did not provide for a good indicator of quality in their results, and we rewrote the questions and answers to improve clarity. Of the 100 generated questions rated by each assessor, 30 questions were the same for each assessor, while the other 70 were unique questions. This enabled us to estimate inter-rater reliability, while still rating a large number of questions overall. Of these 310 unique questions, 155 are questions that the QA filtering model accepted, while the other 155 are questions that the QA filtering model rejected. This should highlight the effect of the QA filtering model and show whether it is a good measure of the quality of questions. 10 example questions used as part of the evaluation are shown in Figure \ref{fig:samples}.

\begin{table*}[t]%% {table*}[t] %%
\centering
\begin{tabular}{llllllll}
\toprule
& & Accepted & Rejected\\
\midrule
\multirow{3}{*}{\small{\textbf{Question 1 (question quality)}}} & \small{Well-formed and understandable} & 70\% & 69\%\\
& \small{Only understandable} & 18\% & 14\%\\
& \small{Neither} & 12\% & 18\%\\
\midrule
\multirow{3}{*}{\small{\textbf{Question 2 (answer compatibility)}}} & \small{Yes} & 50\% & 56\%\\
& \small{No} & 41\% & 37\%\\
& \small{I don't know} & 8\% & 7\%\\
\bottomrule
\end{tabular}
\caption{Results from the human evaluation. We compare the quality of the questions which were accepted by the QA filtering model with those which were rejected.}
\label{table:he_results}
\end{table*}

We estimated the inter-rater reliability of the data using the Fleiss' kappa measure \citep{fleiss1971measuring}. This led to a $\kappa$ value of 0.413 for question 1 and a $\kappa$ value of -0.147 for question 2. Using the interpretation table\footnote{It should be noted that there is extensive debate about the validity of these ranges of interpretation, but it seems to be the most commonly used.} from \citet{landis1977measurement}, the assessors would appear to be in moderate agreement for question 1, but in slight disagreement for question 2. Note that we did not filter the results from question 2 for when the assessors chose the `Neither' option in question 1, which might have influenced the results if the assessors misunderstood the question.

The output of the human evaluation can be found in Table \ref{table:he_results}. The questions which the QA filtering model accepted are overall 5\% point better than those it rejected. 88\% of accepted questions are either only understandable (18\%) or are both well-formed and understandable (70\%). This is a bit higher than the 83\% for rejected questions. However, this is still a pretty small difference. A Pearson’s chi-squared test indicated that the difference between the accepted and rejected questions was not significant ($p=0.21$ for question 1 and $p=0.40$ for question 2).

\section{Discussion}
The results show that the text quality of generated distractors is substantially higher than previous works, that using a larger model has a small effect on the question answering ability, and that the difference in quality when applying the QA filtering model is statistically insignificant, as evaluated by humans. To put these results into context, we need to be aware of the limitations of the different evaluation methods. As for the text generation quality measures such as BLEU and ROUGE, the main issue is that they do not consider the meaning of the text, but only literal word overlap. There is some recent work in using neural language models for evaluating the text quality \citep{sellam-etal-2020-bleurt}, which should better incorporate meaning into the score. This could be considered for future work in this area. Moreover, these metrics do not evaluate sentence structure as part of their calculation.

As for the question answering ability, the main issue is that the model can accept bad questions or reject good questions. These types of errors are included in the total score. Ideally, we would need a QA model which always answers a good question correctly and always answers a bad question incorrectly.
This means that the absolute values from Table \ref{table:qa_ability} contain some noise, but they do give a general indication of relative quality.

As for the human evaluation, the main issue is the low number of total assessed questions, leading to a lack of statistical power. Since there is some subjectivity in how the generated questions are rated by the assessors, we would say that moderate agreement for question 1 is a positive result. The low score for question 2 can be explained by a combination of the question being even more subjective, as well as the fact that question 2 was perhaps not explained well in the evaluation setup. Therefore, we focus primarily on the results for question 1.

The known limitations of evaluation metrics for text generation have led us to use three different evaluation methods. The combined results suggest that whether the question is answerable by the multiple choice QA model, is only a minor indicator of question quality. There was only a small difference in the quantitative results and no statistically significant difference in the qualitative results. One possible reason for this result is that the QA model will guess one of the four options if it does not know the answer for certain, leading to a high false positive rate. This could potentially be resolved by using bayesian neural networks to determine the QA uncertainy and set a threshold, ensuring that the model is sure about its prediction. Or a fifth ``I don't know" option could be added to the QA output and we could teach the model to choose this option when it is not certain.

\section{Conclusions}
Overall, we can conclude that distractor generation using GPT-2 works well: the proposed method beats the state-of-the-art baselines on all BLEU metrics. In addition, we proposed question filtering using a multiple choice QA model. This additional step does not give a significant improvement in the human-experienced question quality.

We have a number of suggestions for future research. First, besides being applied to our own question generator, we could apply our QA filtering model to improve the results of other question generation models. Second, larger pretrained Transformer-based language models could be experimented with on this task. It would be interesting to see how much of an improvement such larger pretrained models could bring.

Specifically, in the near future, we plan to improve the distractor generation model by setting up an end-to-end training pipeline with the question answering model. Inspired by \citet{klein2019learning}, the idea is to generate distractors for a question, then feed this to the QA model, and backpropagate the loss of the QA model with regards to the weights of the DG model. This way, we could teach the DG model to generate distractors such that the QA model could still correctly identify the correct answer, as the current DG model does not have enough inductive bias to generate distractors which are actually incorrect answers.

In summary, we have shown that generating multiple choice questions with distractors is technically possible using Transformer-based language models. This opens up many new possibilities and interesting applications. For example, it could be used to assist teachers in creating multiple choice exams. Or it could be used to automatically quiz students when they are learning. These developments are getting closer to reality and we aimed for this work to provide a valuable contribution towards this hopeful future.

\section*{Acknowledgements}
This work was performed using resources provided by the Academic Leiden Interdisciplinary Cluster Environment (ALICE).

\bibliography{anthology,acl2020}
\bibliographystyle{acl_natbib}

\end{document}